\documentclass[10pt,twocolumn,letterpaper]{article}

\usepackage[pagenumbers]{cvpr} 

\definecolor{cvprblue}{rgb}{0.21,0.49,0.74}
\usepackage[pagebackref,breaklinks,colorlinks,allcolors=cvprblue]{hyperref}

\def\paperID{38} 
\def\confName{CVPR}
\def\confYear{2025}

\title{X-DECODE: EXtreme Deblurring with Curriculum Optimization and Domain Equalization}

\author{
    Sushant Gautam\\
    Mississippi State University\\
    Starkville, MS, USA\\
    {\tt\small sg2223@msstate.edu}
    \and
    Jingdao Chen\\
    Mississippi State University\\
    Starkville, MS, USA\\
    {\tt\small chenjingdao@cse.msstate.edu}
}

\begin{document}
\nocite{*}

\maketitle
\begin{abstract}
Restoring severely blurred images remains a significant challenge in computer vision, impacting applications in autonomous driving, medical imaging, and photography. This paper introduces a novel training strategy based on curriculum learning to improve the robustness of deep learning models for extreme image deblurring. Unlike conventional approaches that train on only low to moderate blur levels, our method progressively increases the difficulty by introducing images with higher blur severity over time, allowing the model to adapt incrementally. Additionally, we integrate perceptual loss and hinge loss during training to enhance fine detail restoration and improve training stability. We experimented with a variety of curriculum learning strategies and explored the impact of train-test domain gap on the deblurring performance. Experimental results on the Extreme-GoPro dataset showed that our method outperforms the next best method by 14\% in SSIM, whereas experiments on the Extreme-KITTI dataset showed that our method outperforms the next best by 18\% in SSIM.
Ablation studies showed that a linear curriculum progression outperforms step-wise, sigmoid, and exponential progressions, while hyperparameter settings such as the training blur percentage and loss function formulation all play important roles in addressing extreme blur artifacts. Datasets and code are available here: \href{https://github.com/RAPTOR-MSSTATE/XDECODE}{https://github.com/RAPTOR-MSSTATE/XDECODE}.
\end{abstract}    
\section{Introduction}
\label{sec:intro}

Restoring blurred images is a fundamental problem in computer vision, with widespread applications in autonomous driving, medical diagnostics, and consumer photography. Motion blur, lens imperfections, and environmental factors introduce severe degradation, making it challenging for traditional restoration techniques to recover lost details. Deep learning-based deblurring models have demonstrated remarkable progress, but they often struggle with extreme levels of blur due to limitations in training data distribution and generalization capabilities.

\begin{figure}[h!tbp]
    \centering
    \includegraphics[width=0.5\textwidth, height=5cm, keepaspectratio]{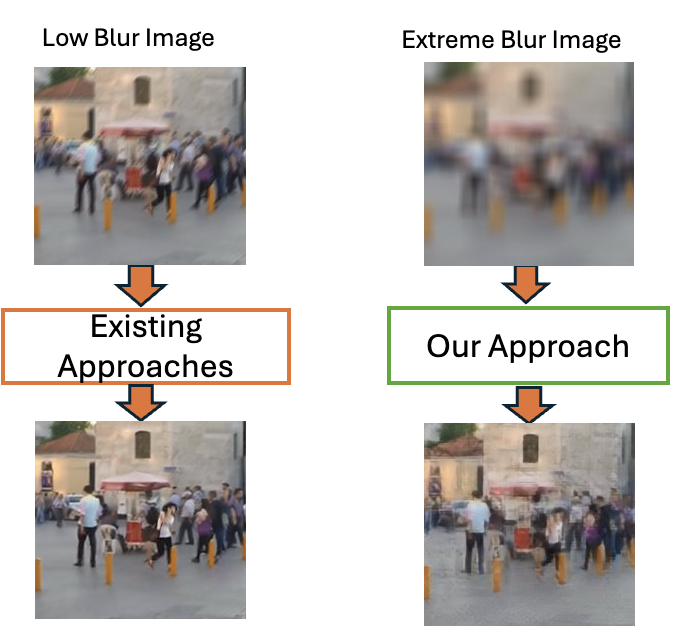}
    \caption{Current deblurring methods achieve effective results only for low-to-moderate blur levels but struggle with extreme blur cases. Our approach bridges this gap by using curriculum learning to train a model to handle extreme degradations.}
    \label{fig:research_gap}
\end{figure}

Current methods typically train models on datasets with a fixed blur distribution, leading to suboptimal performance when encountering significantly degraded images. Conventional techniques such as DeblurGAN \cite{kupyn2018deblurgan}, NAFNet \cite{chen2022simple}, DeblurGAN-v2 \cite{kupyn2019deblurgan}, and AirNet \cite{li2022all} have shown promising results for low or moderate blur levels but exhibit limitations when generalizing to unseen levels of extreme blur. These methods are not explicitly designed to handle extreme blur, leaving a critical gap in the literature (Figure \ref{fig:research_gap}). In particular, existing deblurring models perform deblurring in the general sense without considering the difficulty of removing highly varied levels of blur.

To bridge this gap, the proposed approach leverages curriculum learning \cite{bengio2009curriculum} — an adaptive training strategy inspired by the human learning process, where simpler tasks are introduced before progressively increasing the difficulty level. By gradually training the deblurring model on data with increasing levels of blur severity, we aim to enable the model to adapt incrementally to more challenging restoration tasks. We propose that this systematic training process not only improves robustness but also enhances the model's generalization capabilities across varying levels of image degradation and performs well even when tested on different domains. Our training strategy is architecture-agnostic and can be applied to various generative models, including GAN-based and Transformer-based networks. This work is specifically targeted to the problem of extreme blur artifacts but we expect that the underlying strategies can be extended to other types of extreme image artifacts as well. 
 
In summary, the key contributions of this research are:
\begin{itemize}
    \item A novel curriculum learning strategy for deblurring that increases blur severity according to a specific progression during training.
    \item A hybrid loss function combining perceptual loss, L1 loss, and hinge loss to enhance fine detail preservation and improve training stability.
    \item Extensive experimental evaluation on images with high to extremely high blur levels.
    \item Experimental studies with a variety of curriculum learning progressions, and multiple training and test domains. 
\end{itemize}

\section{RELATED WORK}

\label{sec:relatedwork}

\subsection{Generative Adversarial Networks}
Generative Adversarial Networks (GANs)~\cite{goodfellow2014generative, isola2017image} is a popular approach for image generation via adversarial training. In many paired image-to-image translation tasks, including image restoration, conditional GANs can be used (e.g. pix2pix~\cite{isola2017image}) to generate an image with desired image characteristics conditioned on a source image acting as an input.
Common network architectures for GANs include U-Net-like architectures~\cite{ronneberger2015unetconvolutionalnetworksbiomedical, isola2017image} with skip connections in earlier work, and Transformer-based architectures \cite{zhang2022styleswin, liang2021swinir} in later work. 


\subsection{Artifact Restoration}
Deep learning models for image artifact restoration has advanced quickly in recent years, with all-in-one models such as NAFNet~\cite{chen2022simple} and AirNet~\cite{li2022all}, as well as language-integrated models such as DA-CLIP~\cite{luo2023controlling} and PromptIR~\cite{potlapalli2024promptir}. Multi-stage models such as MPRNet~\cite{zamir2021multi} refine images in sequential steps, while MIRNet~\cite{zamir2020learning} enhances feature extraction for artifact suppression. Transformer-based methods like Restormer~\cite{zamir2022restormer} leverage long-range dependencies for high-resolution restoration. Self-supervised learning, such as Noise2Noise~\cite{lehtinen2018noise2noise}, eliminates the need for clean data, enabling training on corrupted images. However, most existing models are trained on low to moderate levels of image artifacts, leading to a major gap in restoring heavily degraded images.

\subsection{Deblurring Models}
DeblurGAN~\cite{kupyn2018deblurgan} and DeblurGANv2~\cite{kupyn2019deblurgan} are pioneering works that aim to address motion blur using adversarial learning.  Later, Transformer-based models, such as BSSTNet~\cite{zhang2024blurawarespatiotemporalsparsetransformer}, introduce blur-aware attention mechanisms for video deblurring, while event-based methods like EFNet~\cite{sun2022event} integrate event cameras for enhanced motion restoration. Data augmentation strategies, including ID-Blau~\cite{wu2024id}, leverage diffusion-based deblurring to improve generalization across blur levels. However, most of these methods lack specific training mechanisms to adapt to severe blur.

\subsection{Curriculum Learning}
Curriculum learning~\cite{bengio2009curriculum} is a method to improve model training by progressively increasing task complexity. The concept of curriculum learning has been widely applied in diverse scenarios with challenging learning tasks. For example, Oksuz et al.~\cite{oksuz2019automatic} applied this strategy to cardiac MR motion artifact detection, while Gong et al.~\cite{gong2016multi} leveraged multi-modal curriculum learning for semi-supervised classification. However, to the best of our knowledge, curriculum learning has not yet been applied to the problem of severe artifact mitigation in images.

\textbf{Our work} integrates curriculum learning with deblurring models, progressively exposing the network to increasing blur levels during training. This enables better generalization across varying extreme blur severities, thus addressing the limitations of conventional deblurring frameworks.

\section{METHODOLOGY}
\label{sec:methodology}
%

\subsection{Overview}

\begin{figure}[h]
    \centering
    \includegraphics[width=0.5\textwidth]{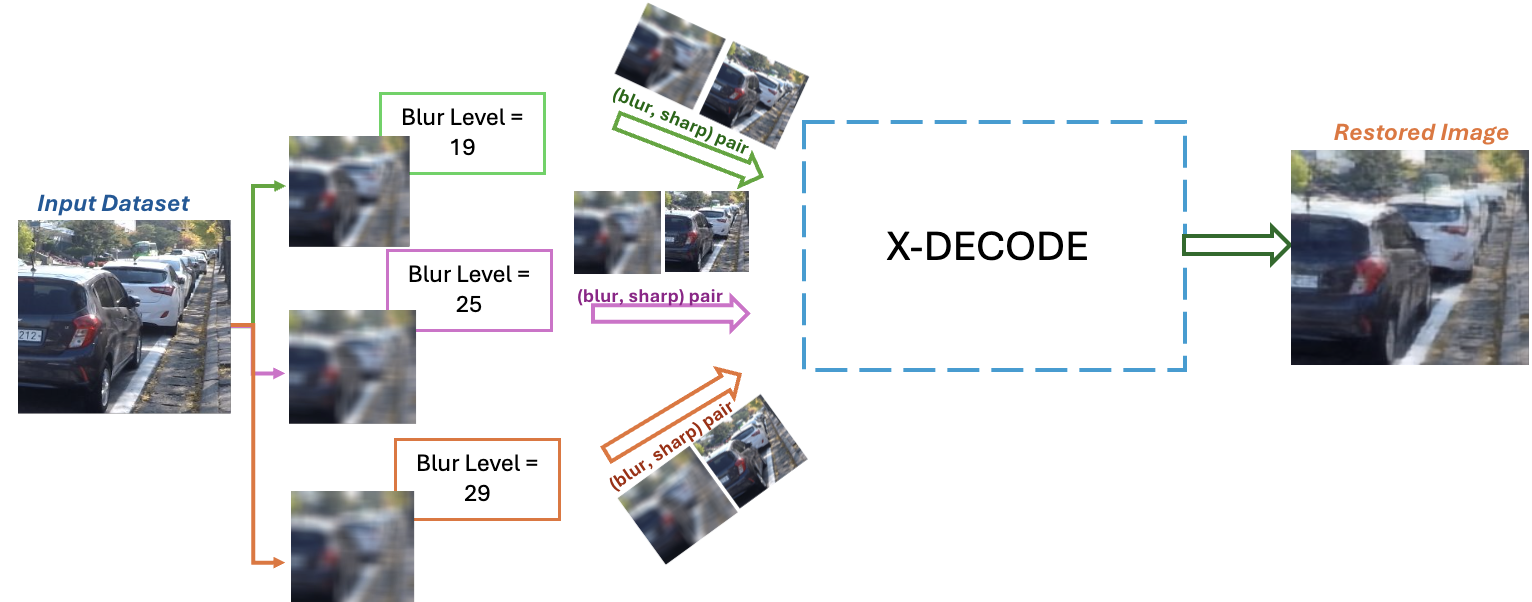}
    \caption{Proposed training strategy for X-DECODE with controlled varying of blur levels}
    \label{fig:methodology_overview}
\end{figure}

\textbf{Figure \ref{fig:methodology_overview}} shows the proposed training strategy, where sharp images from the input dataset are transformed into different blur levels via synthetic blur generation. For each blur level, paired sets of blurred and sharp images are created, and the deblurring model is trained to reconstruct the sharp image from the blurred image via a supervised learning process. The synthetic blur generation process allows the amount of training data to be arbitrarily scalable and the blur level to be easily altered between low blur levels to high blur levels.

\subsection{Model Architecture}
This study utilizes a Generative Adversarial Network (GAN) with a U-net architecture similar to \cite{isola2017image}. However, we intend that the learning techniques presented in this paper are not specific to this architecture and can be easily applied to more advanced transformer-based architectures or backbones.

\subsection{Loss Functions}
In this research, we employ a combination of perceptual loss, L1 loss, and hinge loss to enhance the quality of restored images from high blur levels.

\textbf{1. Perceptual Loss}: Similar to DeblurGAN \cite{kupyn2018deblurgan}, we utilize perceptual loss, in which a pre-trained VGG16 network is used to compare high-level features between the generated and real images. By measuring differences in the VGG feature space, perceptual loss ensures that the generated images maintain structural and textural similarity to the real images, resulting in visually more realistic outputs. Mathematically, the perceptual loss between the generated image \( \hat{y} \) and the real image \( y \) is defined as:
\[
\mathcal{L}_{\text{perc}} = \| \phi((\hat{y})) - \phi(y))) \|_1
\]
where \( \phi \) represents the VGG16 feature extractor. We scale the images to a range of [0, 1] in order to adapt the images to the format expected by VGG16. This loss helps the model focus on perceptual quality, crucial for restoring object appearance under extreme blur.

\textbf{2. L1 Loss}: The L1 loss is a pixel-wise loss that calculates the mean absolute difference between each pixel in the generated image and the real image. It is expressed as:
\[
\mathcal{L}_{\text{L1}} = \| \hat{y} - y \|_1
\]
L1 loss ensures pixel-level accuracy in restored images, complementing perceptual loss by encouraging similarity in fine image details.

\textbf{3. Hinge Loss}: Hinge loss is a popular alternative to the traditional adversarial loss in GAN training, offering improved stability and enhanced image generation quality. It has been successfully used in architectures like Self-Attention GAN (SAGAN) \cite{zhang2019self} and Spectral Normalization GAN (SNGAN) \cite{miyato2018spectral}. The hinge loss modifies the adversarial objective to focus on maximizing the margin between real and generated data, thereby addressing issues like vanishing gradients commonly encountered with binary cross-entropy loss.

The hinge loss for the discriminator is defined as:

\[
\begin{aligned}
L_D ={} & -\mathbb{E}_{x \sim p_{\text{data}}}[\min(0, -1 + D(x))] \\
       & -\mathbb{E}_{z \sim p_z}[\min(0, -1 - D(G(z)))]
\end{aligned}
\]

For the generator, the loss is:

\[ L_G = -\mathbb{E}_{z \sim p_z}[D(G(z))] \]

In these equations, \( D(x) \) represents the discriminator's output for real data \( x \), and \( D(G(z)) \) is the output for generated data \( G(z) \). The hinge loss encourages the discriminator to assign higher scores to real data and lower scores to generated data, effectively creating a margin that the generator must overcome to produce realistic outputs.



The hinge loss approach provides several advantages:
\begin{itemize}
    \item \textbf{Improved Training Stability}: By focusing on margin maximization, hinge loss avoids issues like vanishing gradients, leading to smoother and more reliable training.
    \item \textbf{Enhanced Image Quality}: The emphasis on a clear margin helps the generator produce images with higher realism and diversity.
    \item \textbf{Faster Convergence}: The discriminator's strong feedback mechanism accelerates the training process.
\end{itemize}


\textbf{Total Loss Function}: The total generator loss combines perceptual, L1, and hinge losses as follows:
\[
\mathcal{L}_{\text{total}} = \lambda_{\text{perc}} \mathcal{L}_{\text{perc}} + \lambda_{\text{L1}} \mathcal{L}_{\text{L1}} + \lambda_{\text{G}} \mathcal{L}_{G}
\]
where \( \lambda_{\text{perc}} \), \( \lambda_{\text{L1}} \), and \( \lambda_{\text{G}} \) are weights that control the influence of each component in achieving an optimal balance between perceptual quality and pixel-wise accuracy.

This combination of loss functions allows the model to learn fine details while maintaining stability during training, resulting in improved performance on severely blurred images. The generator and discriminator work together in this framework to enhance image restoration quality, addressing the challenges of extreme artifact degradation. 

\subsection{Training Procedure with Curriculum Learning}
Our training process employs curriculum learning, where the model is progressively trained on image samples with increasing levels of blur severity. This incremental approach helps the model adapt to simpler deblurring tasks before tackling more complex ones, thereby improving its robustness.

During training, the sharp input images are progressively blurred using the \texttt{Blur()} function from the \textit{Albumentations} library \cite{info11020125} to generate image samples with different levels of blur severity (e.g., 19, 21,..., 29). Each blurred image is paired with its corresponding sharp image to form an input-output pair for training. Here, \textbf{Blur Level} (\textbf{BL}) is the kernel size of the uniform box blur, which is constrained to be an odd number in the library implementation.


\begin{figure}[h]
    \centering
    \includegraphics[width=0.5\textwidth]{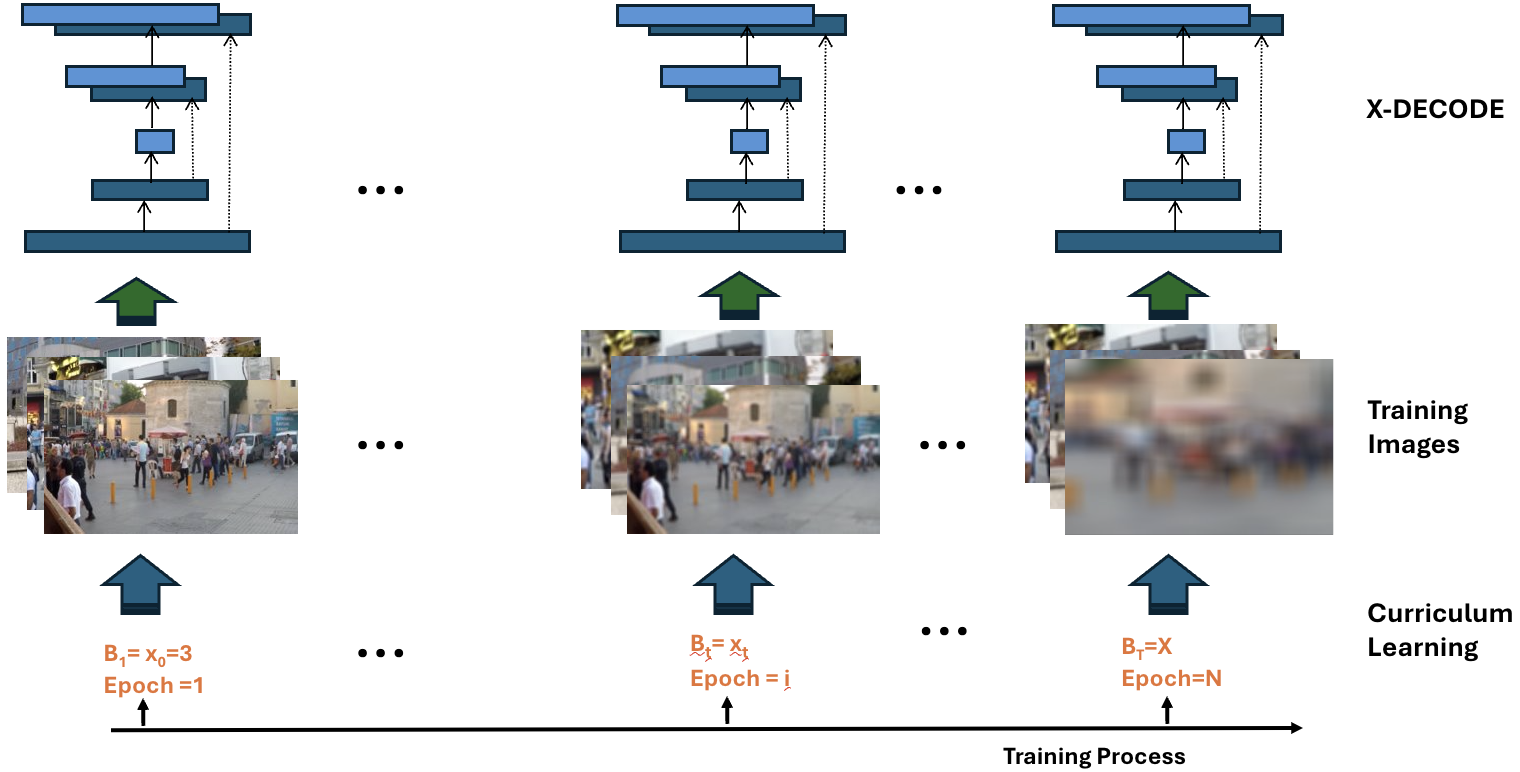}
    \caption{Illustration of a general curriculum strategy for training a deblurring model, where the task difficulty starts with an initial blur level of 3 and increases at every epoch $i$ until the maximum blur level $X$ is reached at $N$ epochs.}
    \label{fig:curriculum_learning_demo}
\end{figure}

\textbf{Figure \ref{fig:curriculum_learning_demo}} illustrates this curriculum learning strategy, which starts with an initial blur level of 3 and gradually increases it in steps, every $i^{th}$ epoch, to reach a maximum blur level of $X$ over the $N$ epochs.




\begin{figure}[h]
    \centering
    \includegraphics[width=0.5\textwidth, height=6cm, keepaspectratio]{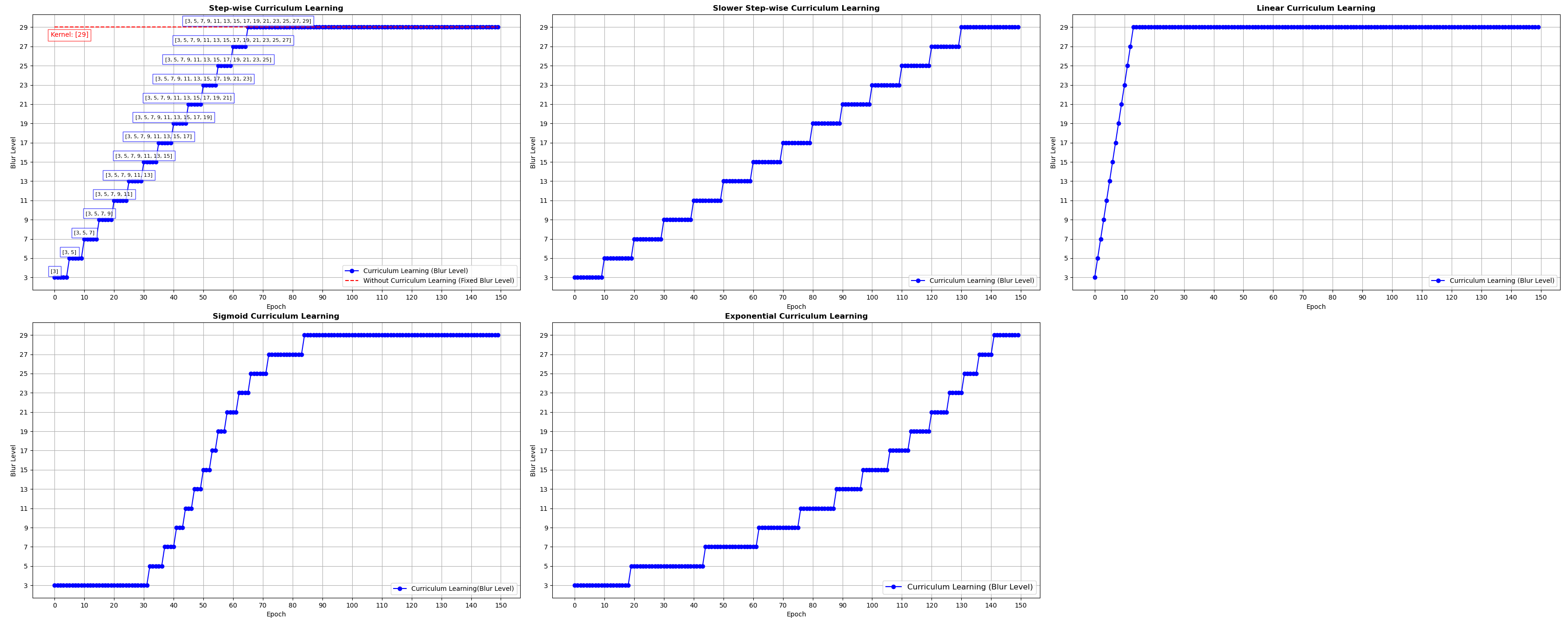}
    \caption{Different blur level progressions for curriculum learning (from left to right, top to bottom: step-wise, slower step-wise, linear, sigmoid, exponential)}
    \label{fig:diff_curr_subplots}
\end{figure}

\subsubsection{Step-wise Curriculum Learning}


In this study, a uniform \textbf{box blur} is applied with a \textbf{blur limit} parameter controlling the blur kernel size, defined as follows:
\begin{equation}
    \text{Blur}(\text{blur\_limit} = (L, L))
\end{equation}
where \( L \) represents the blur level at a given training step. We define \textbf{BL=3} as the minimum blur level and \textbf{BL=29} as the maximum blur level.  We chose \textbf{BL=19} as the starting point at which images can be considered to have "extreme blur", while \textbf{BL=29} is chosen as the upper limit, beyond which additional blur does not significantly change the image appearance further.

Unlike fixed blur training (red dashed line) as shown in Figure ~\ref{fig:diff_curr_subplots}, curriculum learning (blue curve) gradually increases blur severity:
\begin{itemize}
    \item Training starts at \textbf{BL=3} and increases by \textbf{2} every 5 epochs, reaching \textbf{BL=29}.
    \item At each level \( L \), the kernel size is sampled from \(\{3, 5, 7, ..., L\}\), ensuring exposure to varied blur intensities.
\end{itemize}

This step-wise progression helps the model to first encounter mild distortions before handling more extreme distortions, leading to improved training. A similar blur kernel selection process is used in other curriculum learning progressions, as described in the following subsections.

\subsubsection{Slower Step-wise Curriculum Learning}


In slower step-wise curriculum learning, the blur level is increased according to the progression shown in Figure~\ref{fig:diff_curr_subplots}, where the blur level increases every 10th epoch. It is almost the same as the previous step-wise increase approach, but we experiment with how the model performs if the blur level increases slowly over training.                                                 

\subsubsection{Linear Curriculum Learning}


In this experiment, we increase the blur level linearly by 2 every epoch. At epoch 14, it reaches the maximum blur level of 29, as shown in Figure~\ref{fig:diff_curr_subplots}.

\subsubsection{Sigmoid-Based Curriculum Learning}
The sigmoid function is commonly used to model population growth due to its smooth and controlled curve. In curriculum learning, we leverage the sigmoid function to gradually increase the blur level throughout training and then taper it off. Figure~\ref{fig:diff_curr_subplots} illustrates how the blur level progresses over epochs using a sigmoid-based curriculum. The blur level remains low initially, then increases with the highest rate at the midpoint, and finally reaches a plateau at the end. The equation we used is:

\begin{equation}
B_t = \min \left( \max(3, 29 \times \frac{1}{1 + e^{-k(t-m)}}), 29 \right)
\end{equation}
where \( k = 0.1 \) controls the growth rate and \( m = 50 \) defines the midpoint of the transition. 3 is the minimum starting blur level, and 29 is the maximum. 


\subsubsection{Exponential Curriculum Learning}

Exponential growth-based curriculum learning follows a controlled but accelerating schedule, where the blur level increases in a step-wise fashion with an exponential factor as shown in Figure~\ref{fig:diff_curr_subplots}. This ensures a steady rise in difficulty while preventing too rapid an increase. The equation we used is:

\begin{equation}
B_t = \min \left( 2 \times \left\lfloor \frac{B_0 \cdot r^{(t/s)}}{2} \right\rfloor + 1, B_{\max} \right)
\end{equation}
where \( r = 1.15 \) controls the growth rate, \( s = 6 \) determines step intervals, and \( B_0 \) is the initial blur level and \(B_{max}\) is maximum.



Overall, curriculum learning provides a structured way to introduce blur, making training more effective by gradually increasing difficulty while ensuring variability in kernel size selection.

\subsection{Fixed Blur Training Baseline (without Curriculum Learning)}   

In the fixed blur training baseline, represented by the red dashed line as shown in Figure~\ref{fig:diff_curr_subplots}, the \textbf{maximum blur level (29)} is applied consistently throughout the entire training period. The key characteristics of this approach include:

\begin{itemize}
    \item A constant kernel size of \textbf{29×29} is used to blur all images.
    \item The model is immediately exposed to the most challenging blur conditions from the beginning of training.
    \item The lack of progressive difficulty may hinder the model’s ability to generalize effectively, as it does not gradually adapt to increasing blur levels.
\end{itemize}

\section{Experimental Details}
\label{sec:trainingdetails}



\subsection{Hardware and Training Configuration}
Our experiments are conducted on an \textbf{NVIDIA RTX 6000 Ada GPU} (49GB VRAM) and Intel(R) Xeon(R) w5-2465X. Images are resized to \textbf{256×256}, and the model is trained with a \textbf{batch size of 256}.
We experimented with both a fixed learning rate as well as a dynamic learning rate using a scheduler. We also compared the model performance by training (i) with curriculum learning and (ii) without curriculum learning, where the model is trained on fixed blur levels.

\subsection{Dataset}

\textbf{GoPro} is a popular dataset used for training and evaluation of deblurring models \cite{Nah_2017_CVPR}, containing 3,214 blurred images (1280×720 resolution), split into 2,103 training and 1,111 test images. Each blurred image has a corresponding sharp ground truth captured using a high-speed camera.



\begin{figure}[h]
    \centering
    \includegraphics[width=0.5\textwidth]{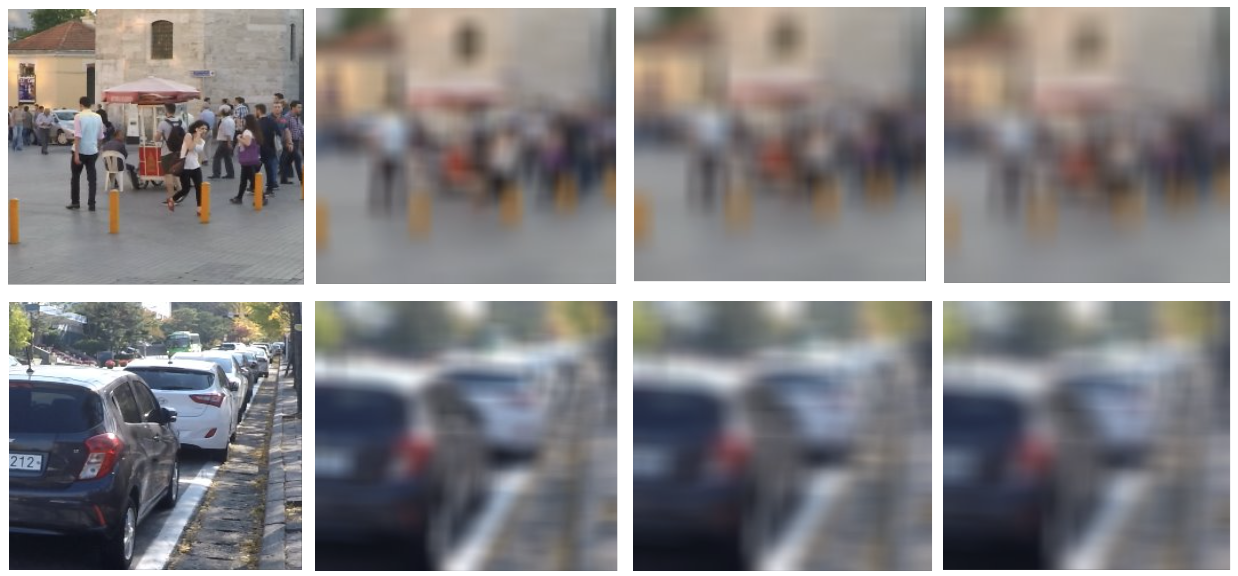}
    \caption{Blur variations in the Extreme-GoPro dataset. Columns show sharp images followed by BL 19, BL 25, and BL 29.}
    \label{fig:diff_blur_levels}
\end{figure}

While GoPro provides high-quality pairs of sharp and blurry images, it has a few limitations which we observed in this study: \textbf{(i)} relatively few number of training samples and \textbf{(ii)} repetitive content due to the images being sampled from the same motion sequences. To investigate generalization, we further incorporate additional datasets: \textbf{Pascal VOC-2012}~\cite{pascal-voc-2012}, \textbf{Cityscapes}~\cite{Cordts2016Cityscapes}, and \textbf{KITTI}~\cite{6248074}. All datasets are resized to a resolution of 256×256 for consistency. KITTI and Cityscapes images are center-cropped to 600×600 before resizing to 256×256 to avoid changing the aspect ratio.

In this study, to simulate extreme blur, we further apply artificial blur levels \textbf{(BL 19 to 29)} to images in the GoPro dataset, creating a new dataset which we call \textbf{Extreme-GoPro}. Similarly, we apply extreme blur levels \textbf{(BL 19 to 29)} to the KITTI dataset, creating a new dataset which we call \textbf{Extreme-KITTI}. For Extreme-GoPro and Extreme-KITTI, we apply blur dynamically during training, but for the test set, the blurry images are generated offline to ensure consistency when comparing performance between different deblurring models.

\begin{figure}[h]
    \centering
    \includegraphics[width=0.5\textwidth]{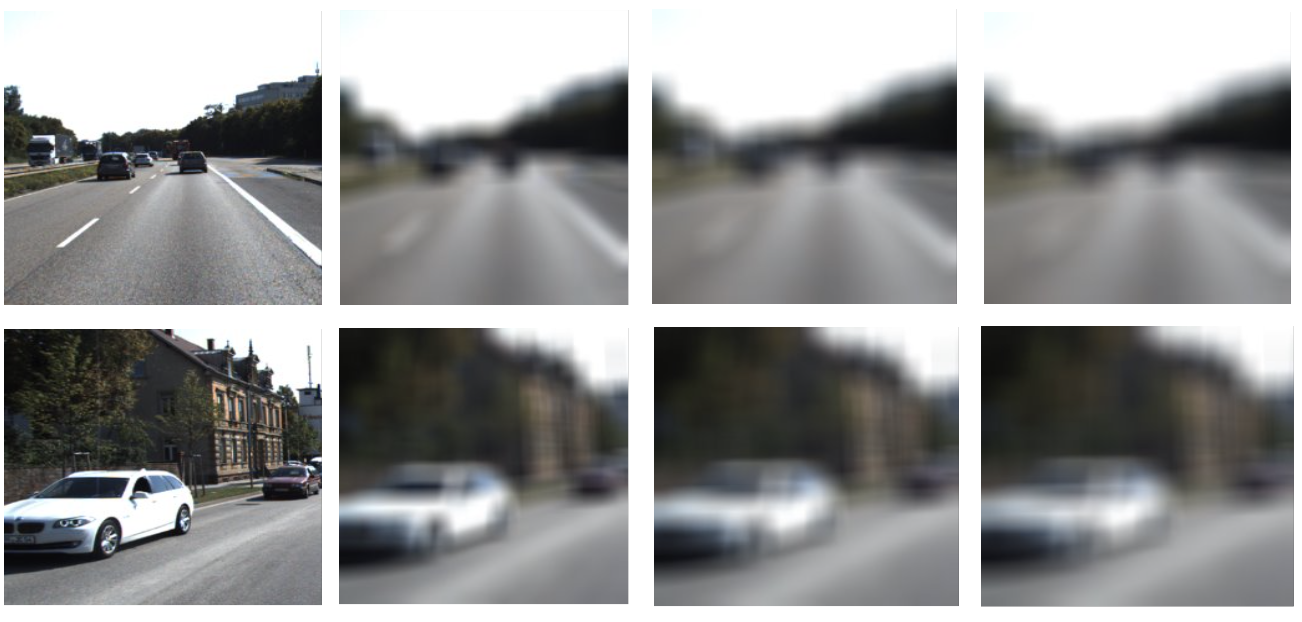}
    \caption{Blur variations in the Extreme-KITTI dataset. Columns show sharp images followed by BL 19, BL 25, and BL 29.}
    \label{fig:diff_blur_levels_kitti}
\end{figure}

Figures \ref{fig:diff_blur_levels} and \ref{fig:diff_blur_levels_kitti} illustrates how increasing blur levels affect image quality in the Extreme-GoPro and Extreme-KITTI datasets.

\subsection{Evaluation Metrics}

To assess the effectiveness of the proposed methodology, we will use standard evaluation metrics such as the peak signal-to-noise ratio (PSNR) and the structural similarity index (SSIM). These metrics will allow a quantitative comparison of the model's performance against baseline models, focusing on improvements in restoration quality and generalization ability. In addition, a qualitative evaluation is conducted by visually comparing restored images to assess perceptual quality.

\section{Result}
\label{sec:results}

\subsection{Deblurring Performance on the Extreme-GoPro Test Set}

To assess the effectiveness of our approach under extreme blur conditions, we evaluate deblurring performance on the Extreme-GoPro test set. Table~\ref{table:deblurring_gopro} provides a quantitative comparison between our method and three baseline models: DeblurGAN~\cite{kupyn2018deblurgan}, DeblurGANv2~\cite{kupyn2019deblurgan}, and NAFNet~\cite{chen2022simple}. The comparison is performed at three different blur levels (\textbf{BL=19, BL=25, BL=29}), with performance reported using the Structural Similarity Index (SSIM) and Peak Signal-to-Noise Ratio (PSNR) metrics. The baseline models are taken from their original published implementations, whereas our model is trained with step-wise curriculum learning.

Results show that our model achieves superior performance at \textbf{BL=19} and \textbf{BL=25}, outperforming the baselines in both SSIM and PSNR. At \textbf{BL=29}, our SSIM score is comparable with the next best method, while our method maintains the best PSNR.

\begin{figure}[h]
    \centering
    \includegraphics[width=\textwidth, height=5cm, keepaspectratio]{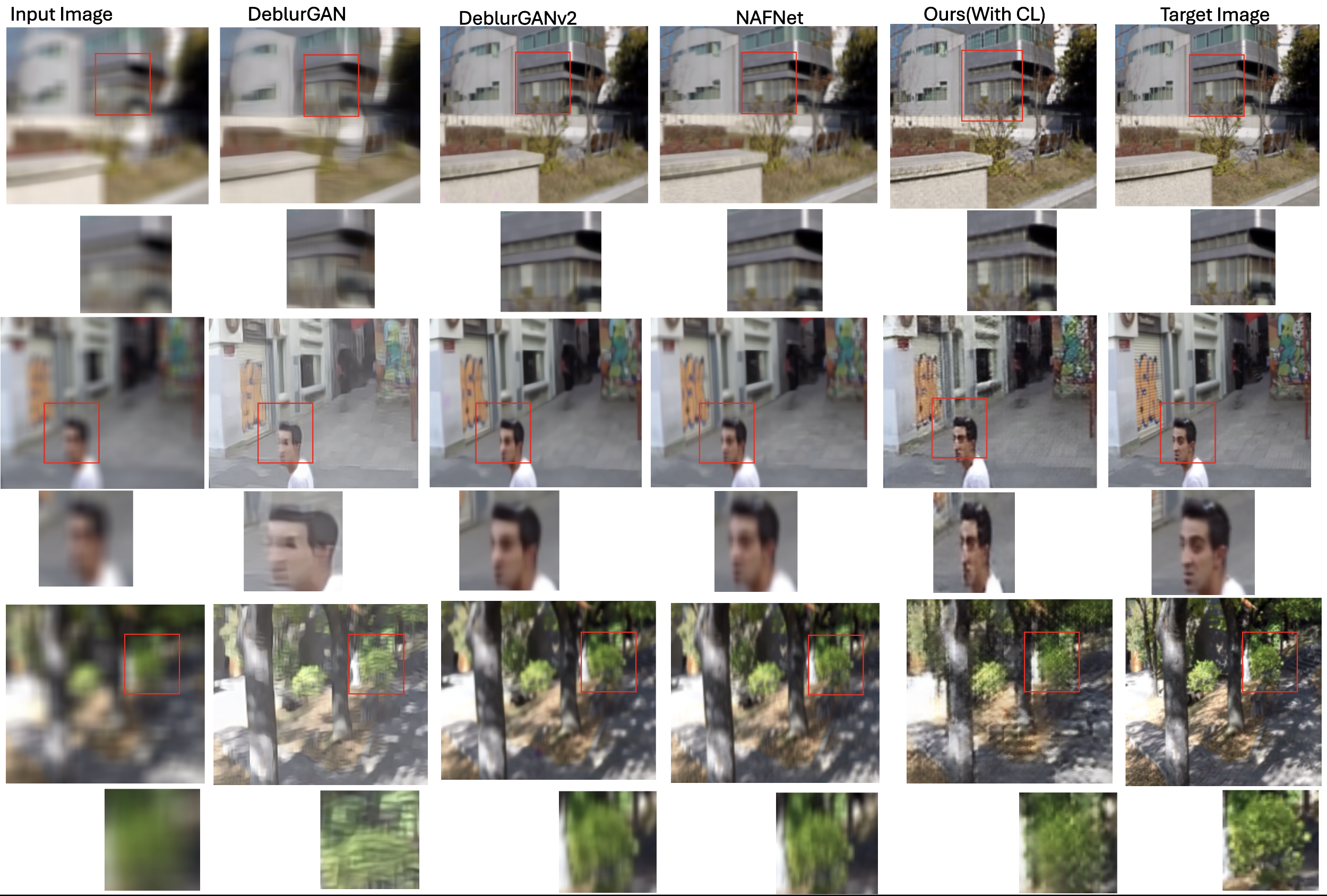}
    \caption{Qualitative evaluation on the Extreme-GoPro dataset with varying blur levels. 
    The rows corresponds to \textbf{BL=19},  \textbf{BL=21}, and \textbf{BL=29}. Each column compares deblurring models, including DeblurGAN~\cite{kupyn2018deblurgan}, DeblurGANv2~\cite{kupyn2019deblurgan}, NAFNet~\cite{chen2022simple}, and our method (with curriculum learning). The rightmost column displays the sharp ground truth images. The red boxes highlight zoomed-in regions for further comparison.}
    \label{fig:gopro-test-comparision-baseline}
\end{figure}

Figure~\ref{fig:gopro-test-comparision-baseline} provides a qualitative comparison of deblurring results at different blur levels. The visual results demonstrate that our method with curriculum learning is more effective in reconstructing fine details, such as textures, facial features, and structural boundaries, than other approaches. 

\begin{table}[h]
\small
  \caption{Comparison of deblurring results on the Extreme-GoPro test set at different blur levels (BL)}
  \label{table:deblurring_gopro}
  \centering
  \begin{tabular}{cccccc}
    \toprule
    BL & Metric & \cite{kupyn2018deblurgan}&\cite{kupyn2019deblurgan}&\cite{chen2022simple}&Ours\\
    \midrule
    \midrule
19 & SSIM & 0.282 & 0.668 & 0.642 & \textbf{0.764} \\
 & PSNR & 19.51 & 23.94 & 22.99 & \textbf{24.90} \\
\midrule
25 & SSIM & 0.258 & 0.635 & 0.577 & \textbf{0.674} \\
 & PSNR & 19.09 & 22.88 & 21.50 & \textbf{23.64} \\
\midrule
29 & SSIM & 0.240 & \textbf{0.618} & 0.545 & \textbf{0.613} \\
 & PSNR & 18.84 & 22.33 & 20.76 & \textbf{22.68} \\
\bottomrule
  \end{tabular}
\end{table} 

\subsection{Deblurring Performance on the Extreme-KITTI Test Set}

To evaluate the generalizability of our proposed approach on other domains, we tested the model on the Extreme-KITTI dataset. Table~\ref{table:deblur_train_gopro_test_kitti} presents a quantitative comparison of various deblurring models, including DeblurGAN~\cite{kupyn2018deblurgan},  DeblurGANv2~\cite{kupyn2019deblurgan}, NAFNet~\cite{chen2022simple}, and our method, both with and without curriculum learning (CL). All models are trained on the GoPro dataset and tested on the Extreme-KITTI dataset, which contain extreme blur images from BL=19 to BL=29. In our approach, linear curriculum learning was used and the blur level was progressively increased from 19 to 29 during the training process.

\begin{figure}[h]
    \centering
    \includegraphics[width=\textwidth, height=5cm, keepaspectratio]{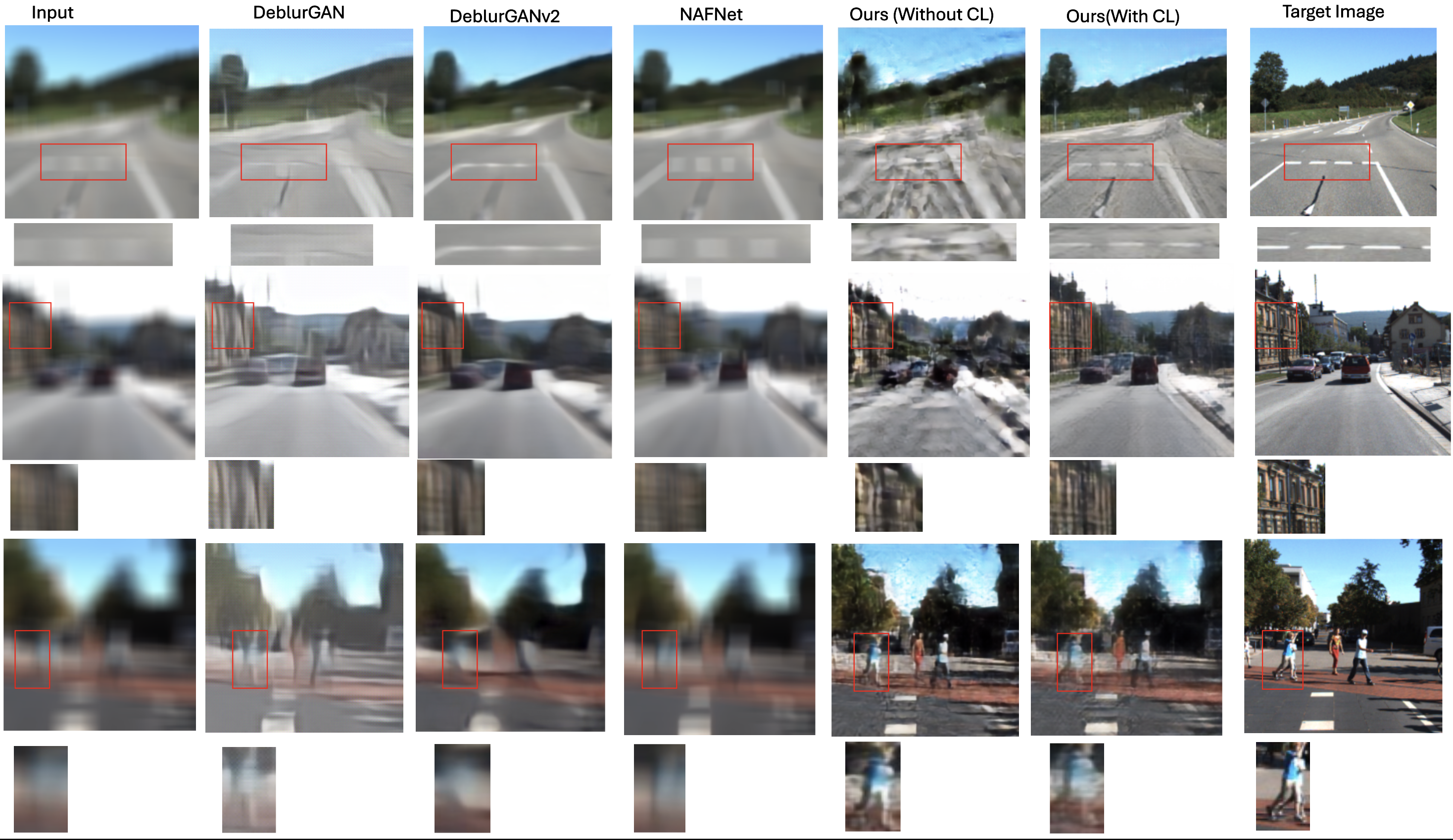}
    \caption{Qualitative evaluation on the Extreme-KITTI dataset with varying blur levels. 
    The rows corresponds to \textbf{BL=19},  \textbf{BL=21}, and \textbf{BL=29}. Each column compares different deblurring models, including DeblurGAN~\cite{kupyn2018deblurgan}, DeblurGANv2~\cite{kupyn2019deblurgan}, NAFNet~\cite{chen2022simple}, and our method (with and without curriculum learning). The rightmost column displays the ground truth sharp images.}
    \label{fig:kitti-test-comparision-baseline}
\end{figure}

Figure~\ref{fig:kitti-test-comparision-baseline} provides a qualitative evaluation of the results across three different blur levels. Notably, our proposed approach with curriculum learning (Ours With CL) produces superior restoration across all blur levels, effectively recovering finer details compared to baseline methods.

The model trained without curriculum learning (Ours Without CL) performs slightly better than the CL approach when tested at the highest blur level (BL=29), as it was solely trained on this blur intensity. However, the curriculum learning approach shows improved generalization at lower blur levels (BL=19 and BL=21), demonstrating its robustness in handling varying degrees of degradation.
In real-world scenarios, it is expected that the blur level of the test images is actually unknown, which also favors our training procedure where the blur levels are varied compared to approaches that train on fixed blur levels.


\begin{table}[h]
  \caption{Comparison of deblurring results on the Extreme-KITTI test set}
  \label{table:deblur_train_gopro_test_kitti}
  \centering
  \begin{tabular}{ccc}
    \toprule
Method & SSIM & PSNR \\
    \midrule
\midrule
DeblurGAN~\cite{kupyn2018deblurgan} &  0.256 & 16.24    \\
DeblurGANv2~\cite{kupyn2019deblurgan} & 0.466  & 18.45   \\
NAFNet~\cite{chen2022simple} & 0.428 & 18.18 \\
Ours (Without CL) &  0.386 & 18.15\\
Ours (With CL) &  \textbf{0.549} & \textbf{20.71} \\ 
\bottomrule
  \end{tabular}
\end{table}

\subsection{Impact of Training Blur Percentage on Deblurring Performance}

Since our experiments indicate that training on easier image reconstruction tasks may be helpful towards learning harder image reconstruction tasks, we hypothesize that inserting a small percentage of sharp-sharp image pairs during training maybe helpful in learning a more generalizable deblurring model. The sharp-sharp image pairs are essentially training samples where the input and output are the exact same sharp image, creating an easy image reconstruction task where the model has to directly learn how to reconstruct a clear image from a latent embedding space. Further, to analyze the effect of different training blur percentages on deblurring performance, we conducted experiments by introducing varying ratios of blurry samples in a training loop. For example, a training blur percentage of 100\% indicates that 100\% of samples in a training loop are sharp-blur image pairs whereas a training blur percentage of 90\% indicates that 90\% of samples in a training loop are sharp-blur image pairs and 10\% of samples in a training loop are sharp-sharp image pairs. Table~\ref{table:blur_percentage} presents the performance on the Extreme-GoPro test set, evaluated at different training blur percentages. The results are reported at three different blur levels (19, 25, and 29), demonstrating the impact at different blur intensities. The models were trained with a combination of L1 loss, hinge loss, and perceptual loss, with weightings of \( \lambda_{\text{L1}} = 1.0 \), \( \lambda_{\text{Perceptual}} = 1.0 \), and \( \lambda_{\text{Hinge}} = 30.0 \).

A key insight from the results is that training with \textbf{100\% ratio of blurred images} results in noticeably lower performance for higher blur levels (\textit{BL 25 and BL 29}), as shown by the reduced PSNR and SSIM values. This suggests that excessive exposure to highly blurred images during training negatively affects the model's ability to learn the relationship between the input and output images.

In contrast, training with \textbf{90\% ratio of blurred images} provides a better balance, leading to improved deblurring results across all blur levels. This suggests that incorporating a small fraction of sharp-sharp image pairs (10\%) during training allows the model to retain the ability to reconstruct essential features of objects despite highly degraded inputs. 

\begin{table}[t]
  \caption{Comparison of training blur percentage on deblurring performance on Extreme-GoPro test set}
  \label{table:blur_percentage}
  \centering
  \begin{tabular}{cccccc}
    \toprule
    Blur level & Metric & 100\% & 95\% & 90\% & 85\%\\
    \midrule
    \midrule
19 & PSNR & 0.726 & 0.723 & 0.764 & 0.766 \\
 & SSIM & 23.68 & 24.26 & 24.90 & 24.98 \\
\midrule
25 & PSNR & 0.223 & 0.656 & 0.674 & 0.535 \\
 & SSIM & 14.75 & 23.35 & 23.64 & 21.40 \\
\midrule
29 & PSNR & 0.169 & 0.627 & 0.613 & 0.525 \\
 & SSIM & 13.29 & 23.07 & 22.68 & 21.21 \\
\bottomrule
  \end{tabular}
\end{table} 


\subsection{Effect of Training Domain}
One of the key goals of our proposed X-DECODE framework is to investigate the ability of deblurring models to generalize across different test domains. We hypothesize that extreme image artifacts may exacerbate the train-test domain gap. However, existing work in deblurring mostly consider cases where the train and test domain are similar, which leaves a major research gap open.




In our experiments in Table~\ref{table:deblurring_diff_dataset}, training on \textbf{Cityscapes} resulted in the highest \textbf{SSIM (0.628)} and \textbf{PSNR (22.20)} scores on the Extreme-KITTI dataset. Whereas, training on the GoPro dataset, which is one of the most widely used datasets for deblurring \cite{kupyn2018deblurgan, kupyn2019deblurgan}, resulted in the lowest scores. We hypothesize that the relatively small number of training samples and repetitive image content in the GoPro dataset limits the ability of a model trained on that dataset to generalize to unseen domains. On the other hand, using the Cityscapes dataset for training enables the model to observe samples of objects similar to that in the KITTI test set (e.g. cars, roads, pedestrians), leading to stronger deblurring performance. In general, our results suggest that deblurring models may rely on learning similar object appearances during training in order to reconstruct them during testing.


\begin{table}[h]
  \caption{Comparison of training domain on cross-domain deblurring performance}
  \label{table:deblurring_diff_dataset}
  \centering
  \begin{tabular}{cccc}
    \toprule
    & \textit{Train on } & & \\
Metric & GoPro & PascalVOC & Cityscapes \\
    \midrule
& \multicolumn{2}{c}{\textit{Test on Extreme-KITTI}} & \\
SSIM & 0.549 & 0.579 &  \textbf{0.628} \\
PSNR & 20.71 & 21.54 & \textbf{22.20} \\
\bottomrule
  \end{tabular}
\end{table} 


\begin{figure}[h]
    \centering
    \includegraphics[width=\textwidth, height=5cm, keepaspectratio]{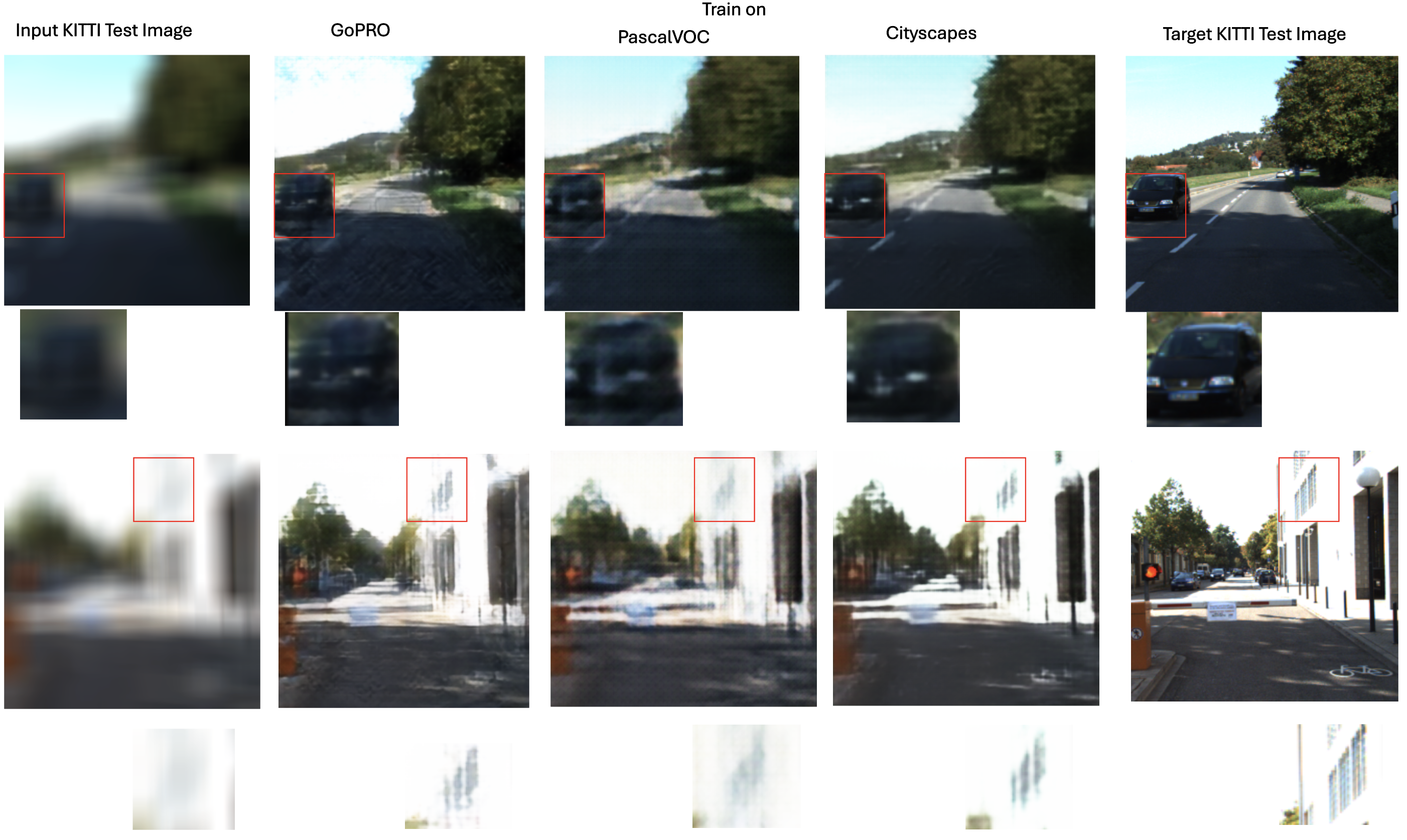}
    \caption{Qualitative comparison of deblurring performance on the KITTI test set when trained on different datasets (GoPro, PascalVOC, and Cityscapes). The leftmost column shows the input blurred KITTI image, followed by the deblurring results from models trained on GoPro, PascalVOC, and Cityscapes. The rightmost column presents the ground truth sharp image. The Cityscapes-trained model shows the best generalization, preserving fine details and structural integrity more effectively.}
    \label{fig:different-test-domains-kitti-comparision}
\end{figure}

\subsection{Comparison of Different Curriculum Learning Progressions}


Table~\ref{table:deblurring_diff_curr_learning} compares different curriculum learning progressions. The models were trained on the \textit{Cityscapes} dataset and tested on the \textit{Extreme-KITTI} dataset, which consists of images with extreme blur levels ranging from 19 to 29. The results show that the \textbf{Linear Curriculum Learning} strategy achieves the highest SSIM (0.628) and PSNR (22.20), suggesting that a gradual, consistent increase in blur severity during training enhances performance. While \textbf{Step-wise} and \textbf{Exponential} approaches also perform well, the \textbf{Slower Step-wise} and \textbf{Sigmoid-based} strategies exhibit lower performance, potentially due to slower adaptation or more abrupt difficulty transitions.



\begin{table}[h]
  \caption{Comparison of different curriculum learning progressions, trained on the Cityscapes dataset and tested on the Extreme-KITTI dataset}
  \label{table:deblurring_diff_curr_learning}
  \centering
  \begin{tabular}{ccc}
    \toprule
Curriculum Learning Progression & SSIM & PSNR \\
    \midrule
\midrule
Step-wise &  0.619 & 21.97    \\
Slower Step-wise & 0.468  & 18.91   \\
Linear & \textbf{0.628} & \textbf{22.20} \\
Sigmoid &  0.405 & 18.02\\
Exponential &  0.614 & 21.79\\
\bottomrule
  \end{tabular}
\end{table}

\section{Ablation Studies}
\label{sec:albationstudies}
To analyze the contribution of each component of our proposed approach, we conducted ablation studies focusing on three key factors: loss function, learning rate scheduling, and mixed-precision training.
or this study, we trained the model on the \textit{Cityscapes} dataset and tested it on the \textit{Extreme-KITTI} dataset, evaluating the deblurring performance across a range of blur severities. We employed the \textbf{Linear Curriculum Learning} strategy to progressively increase blur levels during training.


Table~\ref{table:ablation_studies} compares different loss function configurations, specifically the combination of \textit{L1 Loss, Hinge Loss, and Perceptual Loss}. The results show that the best performance is achieved with \textbf{L1 + Hinge + Perceptual Loss}, outperforming traditional \textbf{L1 + Adversarial Loss} and other combinations.
Adversarial loss is commonly used in the training of GAN-based models. However, we observed that \textbf{Hinge Loss} is more stable and produces higher SSIM and PSNR scores, making it a preferable choice for extreme blur restoration. Additionally, the incorporation of \textbf{Perceptual Loss} further enhances deblurring performance by preserving perceptual quality.

We further evaluated the effect of using a \textbf{Fixed Learning Rate} versus a \textbf{Scheduled Learning Rate}. The results indicate that maintaining a constant learning rate (\textit{lr} = 0.0002) yields superior performance.

Finally, we examined the impact of \textbf{Mixed-Precision Training}, which is known to accelerate computation and reduce GPU memory usage during training. While mixed-precision training results in a minor decrease in performance, the trade-off in training efficiency may be beneficial in certain scenarios.

\begin{table}[h]
  \caption{Ablation Studies}
  \label{table:ablation_studies}
  \centering
  \begin{tabular}{ccc}
    \toprule
Method & SSIM & PSNR \\
    \midrule
\midrule
L1 + Adversarial Loss  &  0.442 & 18.72    \\
L1 + Hinge Loss & 0.621 & 21.91   \\
L1 + Adversarial + Perceptual Loss & 0.266 & 16.15 \\
L1 + Hinge + Perceptual Loss &  \textbf{0.628} & \textbf{22.20}\\
\midrule
Fixed learning rate (lr=0.0002) &  \textbf{0.628} & \textbf{22.20}    \\
Scheduled learning rate & 0.561 & 20.93   \\
\midrule
With mixed-precision training &  0.616 & 21.95    \\
Without mixed-precision training & \textbf{0.628} & \textbf{22.20}   \\
\bottomrule
  \end{tabular}
\end{table} 



\section{Conclusion}
\label{sec:conclusion}
We introduced \textbf{X-DECODE}, a curriculum learning-based approach for extreme image deblurring that enables a model to progressively adapt to increasing blur severity.
Extensive experiments on the \textit{Extreme-GoPro} and \textit{Extreme-KITTI} datasets demonstrate that \textbf{X-DECODE} outperforms conventional deblurring models while improving generalization across domains.
Key findings from our work include the value of incorporating a small fraction of sharp-sharp image pairs during training, and the potential of extreme blur to exacerbate the train-test domain gap.
Future work will explore image restoration of other forms of extreme artifacts, domain adaptation strategies, and architecture considerations for further improvements in extreme deblurring performance.

\section*{ACKNOWLEDGMENT}
The work reported herein was supported by the Intelligence Advanced Research Projects Activity (IARPA) via Department of Interior/ Interior Business Center (DOI/IBC) contract number 140D0423C0075. The U.S. Government is authorized to reproduce and distribute reprints for Governmental purposes notwithstanding any copyright annotation thereon. Disclaimer: The views and conclusions contained herein are those of the authors and should not be interpreted as necessarily representing the official policies or endorsements, either expressed or implied, of IARPA, DOI/IBC, or the U.S. Government.

{
    \small
    \bibliographystyle{ieeenat_fullname}
    \bibliography{main}
}


\end{document}